%% file: main.tex
\documentclass[10pt,twocolumn,letterpaper]{article}

\usepackage{cvpr}              

\input{inputs/preamble}

\definecolor{cvprblue}{rgb}{0.21,0.49,0.74}
\usepackage[pagebackref,breaklinks,colorlinks,citecolor=cvprblue]{hyperref}

\usepackage[ruled,vlined,linesnumbered]{algorithm2e}
\usepackage{algpseudocode}

\def\paperID{9350} 
\def\confName{CVPR}
\def\confYear{2024}

\title{LAA-Net: Localized Artifact Attention Network for Quality-Agnostic and Generalizable Deepfake Detection}

\author{Dat NGUYEN$^{\star}$, Nesryne MEJRI$^{\star}$, Inder Pal SINGH$^{\star}$, Polina KULESHOVA$^{\star}$ \and Marcella ASTRID$^{\star}$, Anis KACEM$^{\star}$, Enjie GHORBEL$^{\star,\rtimes}$, Djamila AOUADA$^{\star}$\\
CVI$^2$, SnT, University of Luxembourg$^{\star}$ \\
Cristal Laboratory, National School of Computer Sciences, University of Manouba$^{\rtimes}$\\
{\tt\small \{dat.nguyen,nesryne.mejri,inder.singh,polina.kuleshova,}\\
{\tt\small marcella.astrid,anis.kacem,enjie.ghorbel,djamila.aouada\}@uni.lu}
}

\begin{document}
\maketitle
\newcount\Comments  
\Comments=1   
\newcommand{\kibitz}[2]{\ifnum\Comments=1\textcolor{#1}{#2}\fi}

\definecolor{darkgreen}{rgb}{0,0.5,0} 
\definecolor{purple}{rgb}{1,0,1} 
\newcommand{\EG}[1]{\kibitz{red}{Enjie: #1}}
\newcommand{\DN}[1]{\kibitz{darkgreen}{Dat: #1}}
\newcommand{\IS}[1]{\kibitz{pink}{Inder: #1}}
\newcommand{\NM}[1]{\kibitz{purple}{Nesryne: #1}}
\newcommand{\DA}[1]{\kibitz{blue}{[DA: #1}]}
\newcommand{\MA}[1]{\kibitz{orange}{Astrid: #1}}
\newcommand{\AK}[1]{\kibitz{cyan}{Anis: #1}}

\newtheorem{Definition}{Definition}

\input{sec/0_abstract}

\input{sec/1_intro}
\input{sec/2_relatedW}
\input{sec/3_method}
\input{sec/4_experiment}

\input{sec/4_x_ablation}

\input{sec/6_conclusion}

\input{sec/X_suppl}

{
    \small
    \bibliographystyle{ieeenat_fullname}
    \bibliography{main}
}


\end{document}

%% file: inputs/preamble.tex
\usepackage[dvipsnames]{xcolor}

\usepackage{epsfig}
\usepackage{caption}
\usepackage{subcaption}
\usepackage{array}
\usepackage{xcolor}
\usepackage{multirow}
\usepackage{multicol}
\usepackage{ragged2e}
\usepackage{arydshln}
\usepackage{indentfirst}
\usepackage{graphicx}
\usepackage{footmisc}
\usepackage[normalem]{ulem}

\makeatletter
\newcommand*\bigcdot{\mathpalette\bigcdot@{1.2}}
\newcommand*\bigcdot@[2]{\mathbin{\vcenter{\hbox{\scalebox{#2}{$\m@th#1\bullet$}}}}}
\makeatother
\DeclareMathOperator*{\argmax}{argmax}

\definecolor{LAA-Net}{HTML}{f27452}
\definecolor{MultiAtt.}{HTML}{4c4cff}
\definecolor{SBI}{HTML}{64ae64}
\definecolor{Xception}{HTML}{a64ca6}
\definecolor{RCCE}{HTML}{a6a6a6}
\definecolor{CADDM}{HTML}{fbde6b}

\definecolor{darkgreen}{rgb}{0,0.5,0} 
\definecolor{purple}{rgb}{1,0,1} 

\newcolumntype{H}{>{\setbox0=\hbox\bgroup}c<{\egroup}@{}}

%% file: sec/0_abstract.tex
\begin{abstract}
This paper introduces a novel approach for high-quality deepfake detection called Localized Artifact Attention Network (LAA-Net). 
Existing methods for high-quality deepfake detection are mainly based on a supervised binary classifier coupled with an implicit attention mechanism. As a result, they do not generalize well to unseen manipulations.
To handle this issue, two main contributions are made. First, an explicit attention mechanism within a multi-task learning framework is proposed. By combining heatmap-based and self-consistency attention strategies, LAA-Net is forced to focus on a few small artifact-prone vulnerable regions. 
Second, an Enhanced Feature Pyramid Network (E-FPN) is proposed as a simple and effective mechanism for spreading discriminative low-level features into the final feature output, with the advantage of limiting redundancy. Experiments performed on several benchmarks show the superiority of our approach in terms of Area Under the Curve (AUC) and Average Precision (AP). The code is available at \url{https://github.com/10Ring/LAA-Net}.
\end{abstract}

%% file: sec/1_intro.tex
\section{Introduction}
\label{sec:intro}
Thanks to the development of generative models, tremendous advances in deepfake creation have been witnessed. Unfortunately, these fake visual data can be employed for malicious purposes, as shown in ~\cite{russia-ukraine-war,african-leaders}. The fact that deepfake generation techniques are rapidly gaining in realism only exacerbates this issue. \textit{It is, therefore, crucial to design methods capable of automatically detecting deepfakes, including the most realistic ones that are commonly referred to as high-quality deepfakes}. Nonetheless, detecting high-quality deepfakes remains extremely challenging as they usually enclose subtle and localized artifacts.

\begin{figure}[t]
    \begin{subfigure}[b]{0.5\textwidth}
        \hspace*{1.1cm}\includegraphics[width=0.7\textwidth]{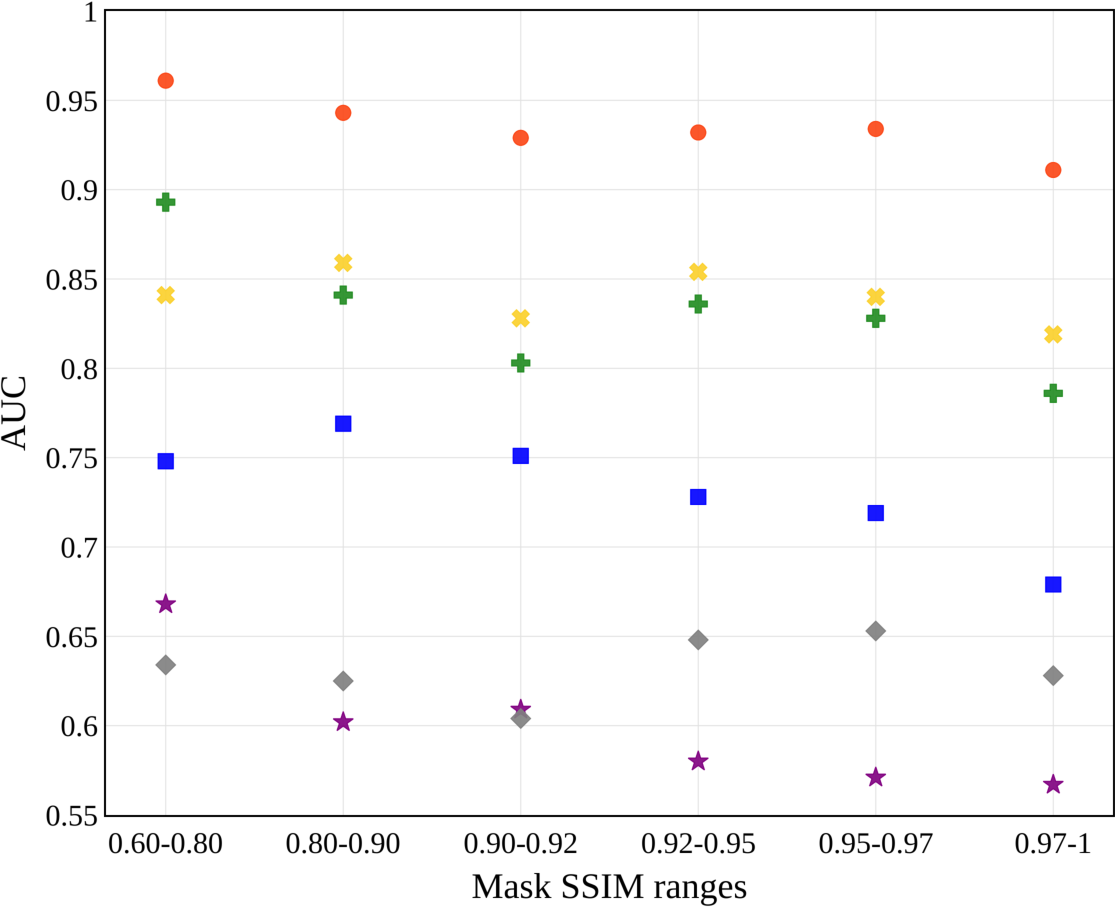}
        \caption{}
    \end{subfigure}
    \begin{subfigure}[b]{0.5\textwidth}
        \hspace*{1.1cm}\includegraphics[width=0.7\textwidth]{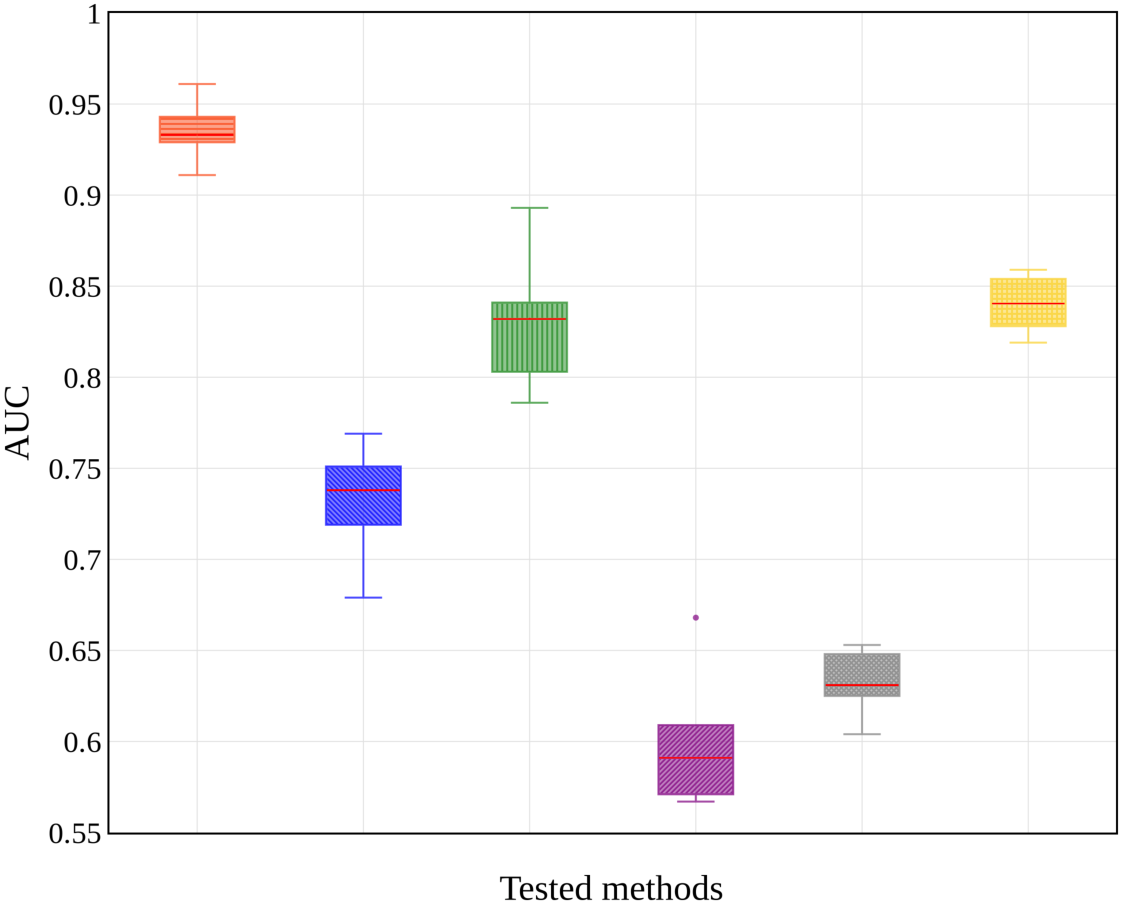}
        \caption{}
        
    \end{subfigure}
    \vspace{-5mm}
    \caption{Comparison of LAA-Net (\textcolor{LAA-Net}{ $\bigcdot$}) with respect to existing methods, namely, Multi-attentional (\textcolor{MultiAtt.}{$\bigcdot$})~\cite{multi-attentional}, SBI (\textcolor{SBI}{$\bigcdot$})~\cite{sbi}, Xception (\textcolor{Xception}{$\bigcdot$})~\cite{ff++}, RECCE (\textcolor{RCCE}{$\bigcdot$})~\cite{ete_recons}, CADDM (\textcolor{CADDM}{$\bigcdot$})~\cite{caddm}, using (a) the AUC performance with respect to different ranges of Mask SSIM, and (b) its associated boxplots.  *The results were obtained using the official source codes pretrained on FF+~\cite{ff++} and testing on Celeb-DFv2~\cite{celeb_df}. Figure best viewed in colors.}
    \vspace{-3mm}
\label{fig:motivation}
\end{figure}

Recent works have mostly focused on improving the generalization capabilities of deepfake detection methods by adopting multi-task learning~\cite{fxray, cstency_learning, sladd} and/or heuristic fake data generation~\cite{fxray, sbi} strategies. However, most of these methods fail to model localized artifacts, which are critical for detecting high-quality deepfakes. This could be explained by the fact that Vanilla Deep Learning (DL) architectures are mainly used. These common architectures, such as XceptionNet~\cite{xception} and EfficientNet~\cite{efn_net}, tend to learn global features, ignoring more localized cues~\cite{multi-attentional, sfdg}. With the use of successive convolutions, localized features across layers gradually fade. Hence, proposing suitable mechanisms for capturing local and subtle artifacts turns out to be necessary.

To the best of our knowledge, only a few research works have explored this research direction~\cite{multi-attentional,sfdg}. They mainly introduce attention modules that implicitly model subtle inconsistencies through low-level representations~\cite{multi-attentional,sfdg}. Nevertheless, they still rely on single binary classifiers trained with real/deepfake images without considering any additional strategy for avoiding generalization issues. This considerably restricts the practical usefulness of these methods.

Hence, our goal is to address the detection of high-quality deepfakes and, at the same time, improve the generalization performance. We argue that this can be achieved by designing an attention module compatible with generic deepfake detection strategies. In particular, the solution would be to introduce an explicit fine-grained mechanism within a multi-task learning framework supported by an appropriate pseudo-fake synthesis technique. Moreover, in addition to such a learning strategy, we posit that an adequate architecture preserving low-level features could implicitly contribute to better capturing localized artifacts.

More concretely, this paper proposes a novel fine-grained approach called \textit{Localized Artifact Attention Network (LAA-Net)} that relies on a multi-task learning framework.
First, a new fine-grained mechanism that aims at focusing on small regions centered at the vulnerable pixels is introduced. By vulnerable pixels, we mean the pixels that are more likely to showcase a blending artifact\footnote{A more formal definition is given in Section~3.2}. This is achieved by considering two auxiliary branches, namely, \textit{a heatmap branch} and a \textit{self-consistency branch}. On the one hand, the heatmap branch allows localizing the set of vulnerable pixels while taking into account their neighborhood. On the other hand, the self-consistency branch estimates the similarity of pixels with respect to a randomly selected vulnerable point. To simulate fake data and generate ground-truth heatmaps and self-consistency matrices that are predicted by the additional branches, blending-based data synthesis such as~\cite{fxray,sbi} are leveraged. Second, the proposed architecture incorporates a novel, simple, yet effective Feature Pyramid Network (FPN)~\cite{fpn_obdet} termed \emph{Enhanced FPN} (E-FPN). It enables making use of multi-scale features while avoiding redundancy. In fact, it has been shown that reducing feature redundancy contributes to the regularization of Deep Neural Networks~(DNNs)~\cite{ayinde2019regularizing}. While the proposed attention mechanism guided by the vulnerable points helps the network to focus explicitly on artifact-prone regions, E-FPN forces the model to consider implicitly local cues. The association of these two complementary components makes LAA-Net a suitable candidate for fine-grained and generic deepfake detection. As reflected in Figure~\ref{fig:motivation}, our approach achieves better and more stable Area Under the Curve (AUC) performance as compared to existing methods~\cite{multi-attentional,sbi,ff++,ete_recons, caddm} regardless of the quality of deepfakes, quantified using the Mask Structural SIMilarity (Mask-SSIM\footnote{The Mask-SSIM~\cite{mssim_pose, celeb_df} has been proposed as a metric for quantifying the quality of deepfakes~\cite{celeb_df}. The Mask-SSIM is computed by computing the similarity in the head region between the fake image and its original version using the SSIM score introduced in~\cite{ssim_1}. Hence, a higher Mask-SSIM score corresponds to a deepfake of higher quality. \label{mssim_defi}}).  
For a more comprehensive evaluation, in addition to the standard AUC, other metric is reported, namely, Average Precision (AP). We report experiments on several deepfake benchmarks and show that LAA-Net outperforms the state-of-the-art (SoA).

\vspace{0.1cm}
\noindent\textbf{Contributions.} In summary, the paper contributions are:
\begin{enumerate}
    \item A novel multi-task learning method for fine-grained and generic deepfake detection called LAA-Net. It is trained using real data only. 
    \item An explicit attention mechanism for focusing on vulnerable points combining heatmap-based and self-consistency attention strategies.
    \item A new FPN design, called E-FPN, ensures the efficient propagation of low-level features without incurring redundancy\footnote{E-FPN is generic and can be used in conjunction with any traditional encoder-decoder architecture.}.
    \item Extensive experiments and a comprehensive analysis reported on several benchmarks, namely, FF++~\cite{ff++}, CDF2~\cite{celeb_df}, DFD~\cite{dfd}, DFDC~\cite{dfdcp}, and DFW~\cite{wdf}.
\end{enumerate}


\vspace{0.1cm}
\noindent\textbf{Paper Organization.}
The remainder of the paper is organized as follows: Section~\ref{sec:related_work} reviews related works. Section~\ref{sec:method} introduces the proposed approach, and Section~\ref{sec:experiment} reports the experiments and discusses the results. Finally, Section~\ref{sec:conclu} concludes this work and suggests future investigations.

%% file: sec/2_relatedW.tex
\begin{figure*}
\begin{center}
\includegraphics[width=0.95\linewidth]{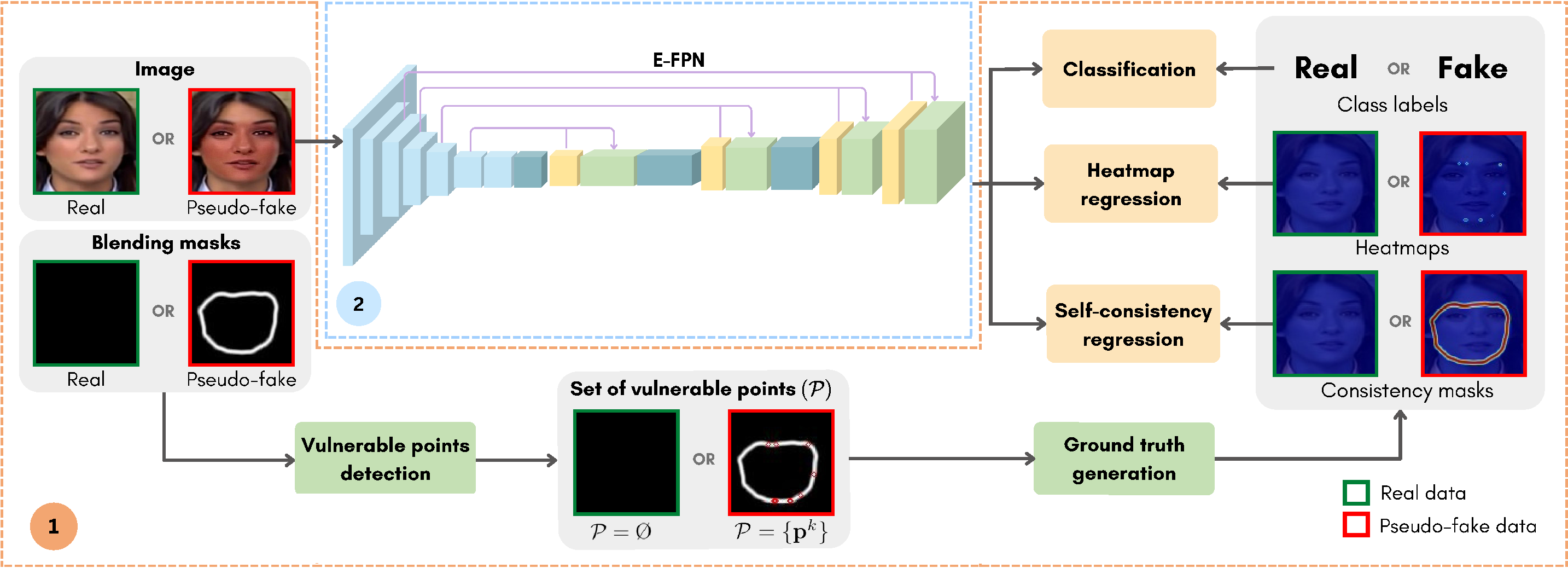}
\vspace{-5mm}
\end{center}
   \caption{Overview of the proposed LAA-Net approach: it is formed by two components, namely, (1) an \textit{explicit attention mechanism} based on a multi-task learning framework composed of three branches, i.e., the binary classification branch, the heatmap branch, and the self-consistency branch. The heatmap and self-consistency ground-truth data are generated based on the detected vulnerable points, and (2) an \textit{Enhanced Feature Pyramid Networks (E-FPN)} that aggregates multi-scale features.}
\vspace{-5mm}
\label{fig:wfl_ovv}
\end{figure*}

\section{Related Works: Attention-based Deepfake Detection}
\label{sec:related_work}

Prior works are diverse in the way they approach the problem of deepfake detection~\cite{vis_artifacts, df_cre_detect, capsulenet, mesonet, fake_spotter, hfc, mlc-inder}. Earlier methods generally formulate it as a purely binary classification~\cite{ff++,efn_vit}, leading to poor generalization capabilities. As a solution, two main strategies have been investigated by the research community, namely, multi-task learning~\cite{fxray, cstency_learning, sladd, mae, leveraging-th, ete_recons} and/or pseudo-fake generation~\cite{untag, fxray, sbi, aunet, altfreezing}.

Despite their great potential, the aforementioned models are less robust when considering high-quality deepfakes.  Indeed, these SoA methods mainly employ traditional DNN backbones such as XceptionNet~\cite{xception} and EfficientNet~\cite{efn_net}. Hence, through their successive convolution layers, they implicitly generate global features. As a result, low-level cues, that can be very informative, might be unintentionally ignored, leading to poor detection performance of high-quality deepfakes. It is, therefore, crucial to design adequate strategies for modeling more localized artifacts.

Alternatively, some attention-based methods such as~\cite{multi-attentional,sfdg} have been proposed. Specifically, they have made attempts to integrate attention modules for implicitly focusing on low-level artifacts~\cite{multi-attentional,sfdg}. Unfortunately, the two aforementioned methods make use of a unique binary classifier solely trained with real and deepfake images. This means that they do not consider any pseudo-fake generation technique or multi-task learning strategy. Consequently, as demonstrated experimentally, they do not generalize well to unseen datasets in comparison to other recent techniques~\cite{sbi, aunet, altfreezing}. 

%% file: sec/3_method.tex
\section{Localized Artifact Attention Network (LAA-Net)}
\label{sec:method}

Our goal is to introduce a method that is robust to high-quality deepfakes yet capable of handling unseen manipulations. Accordingly, we introduce a fine-grained method called Localized Artifact Attention Network (LAA-Net) illustrated in Figure~\ref{fig:wfl_ovv}. LAA-Net incorporates: (1) an \textit{explicit attention mechanism} and (2) a new architecture based on an \textit{enhanced FPN}, called \textit{E-FPN}.

First, the proposed attention mechanism aims at explicitly focusing on blending artifact-prone pixels referred to as vulnerable points (a formal definition is given in Section~\ref{subsec:efag}). For that purpose, a hand-free annotation of vulnerable points is proposed by leveraging a blending-based data synthesis. Specifically, a multi-task learning framework composed of three simultaneously optimized branches, namely (a) classification, (b) heatmap regression, and (c) self-consistency regression, is introduced, as depicted in Figure~\ref{fig:wfl_ovv}. The classification branch predicts whether the input image is fake or real, while the two other branches aim at giving attention to vulnerable pixels.
Second,  E-FPN allows extracting multi-scale features without injecting redundancy. This enables modeling low-level features, which can better discriminate subtle inconsistencies.

\subsection{Explicit Attention to Vulnerable Points}
\label{subsec:efag}

\subsubsection{Blending-based Data Synthesis}
\label{subsec:blending_synthesis}


We start by recalling blending-based data synthesis methods such as~\cite{fxray,sbi}. In fact, the proposed method relies on this kind of pseudo-fake generation and, therefore avoids using actual deepfakes and manually annotating data to train the proposed multi-task learning framework.
Let us consider a manipulated face image denoted by $\mathbf{I_{\text M}}$. The image $\mathbf{I_{\text M}}$ can be obtained by combining (e.g., blending) two images denoted by $\mathbf{I_{\text F}}$ and $\mathbf{I_{\text B}}$ as follows, 

\begin{equation}
    \mathbf I_{\text M} = \mathbf M \odot \mathbf I_{\text F} + (1-\mathbf M) \odot \mathbf I_{\text B} \ , 
    \label{equa:blending_fomular}
\end{equation}
where $\mathbf I_{\text F}$ refers to the foreground image enclosing the desired facial attributes, $\mathbf I_{\text B}$ indicates a background image, $\mathbf M$ is the deformed Convex Hull mask with values varying between $0$ and $1$, and $\odot$ denotes the element-wise multiplication operator.

\subsubsection{Proposed Multi-task Learning Framework }

In addition to the deepfake classification branch, the network learns to focus on specific regions by taking advantage of the parallel \textit{Heatmap} and \textit{Self-consistency} branches. Our hypothesis is that deepfake detection can be formulated as a fine-grained classification. 
Therefore, giving more attention to the vulnerable points should be an effective solution for detecting high-quality deepfakes.
For the sake of clarity, we start by formally defining the notion of \emph{`vulnerable points'}.

\begin{figure}[t]
    \centering
    \includegraphics[width=0.95\linewidth]{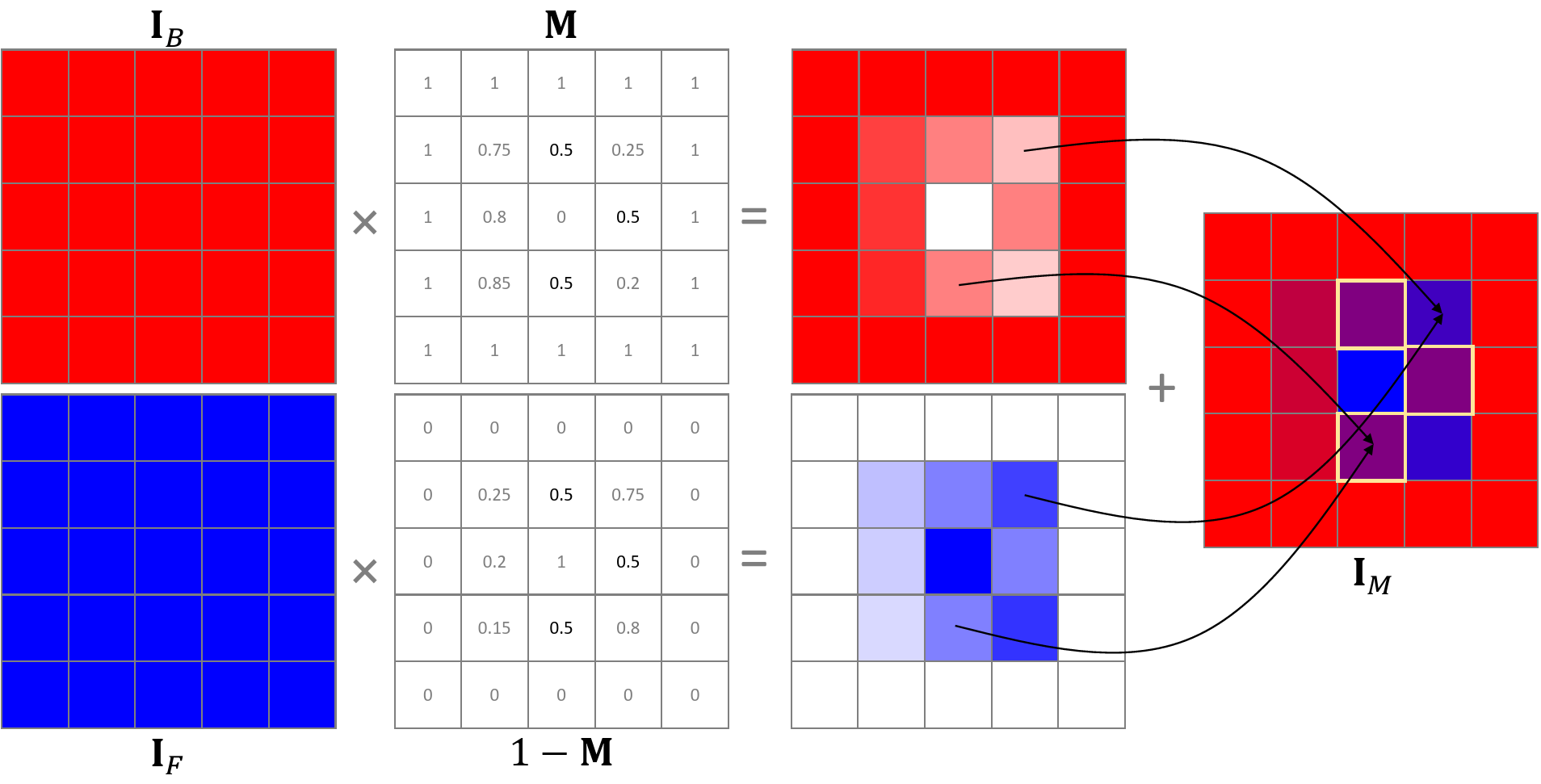}
    \caption{Extraction of the vulnerable points.}
    \label{fig:vulnerable_points_simple}
\end{figure}

\begin{Definition}
 - Vulnerable points in a deepfake image are the pixels that are more likely to carry blending artifacts. 
\end{Definition}
\noindent As discussed in Section~\ref{subsec:blending_synthesis}, any deepfake generation approach involves a blending operation for mixing the background and the foreground of two different images $\mathbf{I}_B$ and $\mathbf{I}_F$, respectively. This implies the presence of blending artifacts regardless of the used generation approach. Thus, we posit that the vulnerable points can be seen as the pixels belonging to the blending regions with the most equivalent contributions from both $\mathbf{I}_B$ and $\mathbf{I}_F$. 

\noindent In this paper, we assume that we work under a realistic setting where we only have access to real data during training. A blending-based augmentation is, therefore, considered and leveraged for defining vulnerable pixels. Specifically, inspired from \cite{fxray}, a blending boundary mask $\mathbf{B}=~(b_{ij})_{i, j \in [\![ 1, D]\!]  }$ is firstly computed as follows, 

\begin{equation}
  \mathbf B = 4\text{ . } \mathbf{M}\odot(\mathbf{1}-\mathbf{M}) \ ,  
\end{equation}
with $\mathbf 1$ being an all-one matrix. Note that $\mathbf{M}$ is defined in ~\cref{equa:blending_fomular}. The variable $D$ is the height and width of $\mathbf B$,  and $b_{ij}$ its value at the position $(i,j)$.  A higher value of $b_{ij}$ indicates that the position $(i,j)$ is more impacted by the blending. Hence, if an input image is real, $\mathbf B$ should be set to $\mathbf 0$. Then, the set of vulnerable pixels denoted by $\mathcal{P} $ is defined as follows,
\begin{equation}
\mathcal P = \argmax_{(i,j) \in [\![ 1,D ]\!] ^2 }(\mathbf{B}) , 
\end{equation}
where $[\![ \textbf{ } ]\!] $ defines an integer interval.
Figure~\ref{fig:vulnerable_points_simple} illustrates the extraction of vulnerable points.
In the following, we describe how the notion of vulnerable points is used within the heatmap and self-consistency branches.

\paragraph{Heatmap Branch.}
\label{subsec:heatmapbranch}
In general, forgery artifacts not only appear in a single pixel but also affect its surroundings. Hence, considering vulnerable points as well as their neighborhood is more appropriate for effectively discriminating deepfakes, especially in the presence of images with local irregularities caused by noise or illumination changes. To model that, we propose to use a heatmap representation that encodes at the same time the information of both vulnerable points as well as their neighbor points. 

More specifically, ground-truth heatmaps are generated by fitting an \textit{Unnormalized Gaussian Distribution} for each pixel  $\mathbf{p}^{k}$ $\in$  $\mathcal P$. The pixel $\mathbf{p}^{k}$  is considered as the center of the Gaussian Mask $\mathbf{G}^{k}$. To take into account the neighborhood information of $\mathbf{p}^{k}$, the standard deviation of $\mathbf{G}^{k}$ is adaptively computed. In particular, inspired from the work of~\cite{cornetnet}, the standard deviation $\sigma_{k}$ of $\mathbf{p}^{k}$ is computed based on the width and the height of the blending boundary mask $\mathbf B$ with respect to the point $\mathbf{p}^{k}$. Similar to \cite{cornetnet}, a radius $r_k$ is computed based on the size of the set of virtual objects that overlap the mask centered at $\mathbf{p}^{k}$ with an Intersection over Union (IoU) greater than a threshold $t$. 
In all our experiments, we set $t$ to 0.7 and we assume that $\sigma_k = \frac{1}{3}r_k$. Hence, $\mathbf{G}^{k}=~(g_{ij}^{k})_{i, j \in [\![ 1, D]\!]  }$ is computed as follows,

\begin{equation}
g_{ij}^{k}= e^{-\frac{i^2+j^2}{2\sigma_{k}^2}} \ , 
\label{equa:2D_unGauss}
\end{equation}
where $i$ and $j$ refer to the pixel position.
The ground-truth heatmap $\mathbf H$ is finally constructed by superimposing the set $\mathcal G= \{ \mathbf{G}^{k}\}_{k \in  [\![ 1, \text{card}(\mathcal P)]\!]}$. A figure depicting the heatmap generation process is provided in supplementary materials.

For optimizing the heatmap branch, the following focal loss~\cite{focal_loss} is used, 
\begin{equation}
 {L}_\text{H}= \sum_{i,j}^{D}{-(1-\Tilde{h}_{ij})^{\gamma}\log{\Tilde{h}_{ij}}} \ ,
\end{equation}
such that,

\begin{equation}
 \Tilde{h}_{ij} =
\begin{cases}
    \hat{h}_{ij} & \text{ if } h_{ij} = 1 \ , \\
    1 - \hat{h}_{ij} &  \text{ otherwise } \ , \\ 
\end{cases}
\end{equation}

\noindent with $\hat{h}_{ij}$ and $h_{ij}$ being the value of the predicted heatmap $\hat{\mathbf{H}}$ and the ground-truth $\mathbf{H}$ at the pixel location $(i,j)$, respectively. The hyperparameter $\gamma$ is used to stabilize the adaptive loss weights.

\paragraph{Self-consistency Branch.} To enhance the proposed attention mechanism, the idea of learning self-consistency proposed in~\cite{cstency_learning} is revisited to fit our context. Instead of computing the consistency values for each pixel of the mask, we consider only the vulnerable location. Since the set $\mathcal P$ might include more than one pixel (the blending mask can include several pixels with equal values), we randomly choose one of them that we denote by $\mathbf p^s$  for generating the self-consistency ground-truth matrix. Hence, the generated matrices denoted by $\mathbf{C}$ are $2$-dimensional and not $4$-dimensional as in the original method. Given the randomly selected vulnerable point $\mathbf p^s=(u,v)$, the self-consistency $\mathbf C$ matrix is computed as, 
\begin{equation}
\mathbf C = \mathbf 1 - |b_{uv}.\mathbf 1 - \mathbf{B}| \ ,
\label{eq:consistency_gt}
\end{equation}
where $|.|$ refers to the element wise modulus and $\mathbf 1$ is an all-one matrix.

This refinement allows for reducing the model size and, consequently, the computational cost. It can also be noted that even though our method is inspired by~\cite{cstency_learning}, our self-consistency branch is inherently different. In~\cite{cstency_learning}, the consistency is calculated between the foreground and background, whereas we measure the consistency between the vulnerable point and the other pixels of the blended mask. 
The self-consistency loss $L_\text{C}$ is then computed as a binary cross entropy loss between $\mathbf C$ and the predicted self-consistency~$\hat{\mathbf C}$.

\paragraph{Training Strategy.} The network is optimized using the following loss,

\begin{equation}
    L = {L}_{\text{BCE}} + \lambda_1 {L}_{\text{H}}  + \lambda_2 {L}_{\text{C}} \ ,
\label{equa:total_loss} 
\end{equation}

\noindent where ${L}_{\text{BCE}}$ denotes the binary cross-entropy classification loss. ${L}_{\text{H}}$ and ${L}_{\text{C}}$ are weighted by the hyperparameters $\lambda_1$ and $\lambda_2$, respectively. Note that only real and pseudo-fakes are used during training.

\subsection{Enhanced Feature Pyramid Network (E-FPN)}
\label{subsec:efpn}

\begin{figure*}[t]
\begin{center}
   \includegraphics[width=0.9\linewidth]{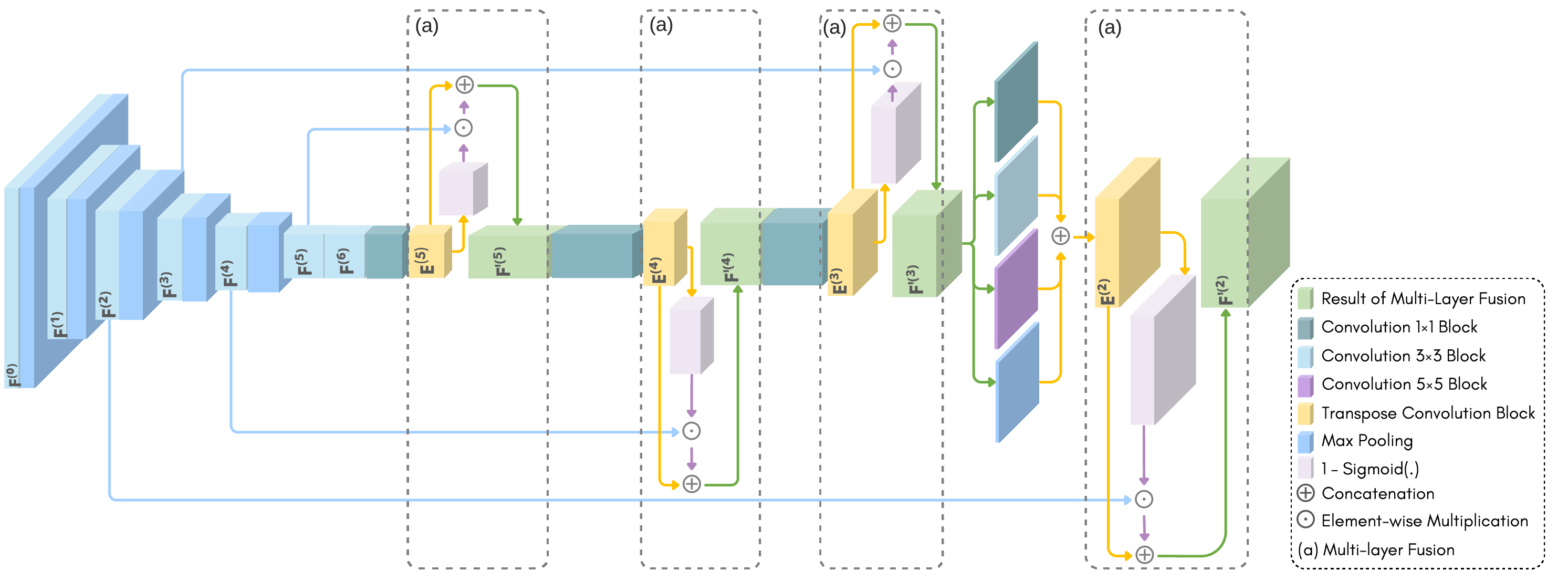}
   \vspace{-5mm}
\end{center}
   \caption{Architecture of the proposed Enhanced Feature Pyramid Network (E-FPN).}
   \vspace{-2mm}
\label{fig:efpn}
\end{figure*}

Feature Pyramid Networks (FPN) are widely adopted feature extractors capable of complementing global representations with multi-scale low-level features captured at different resolutions~\cite{fpn_obdet}. This makes them ideal candidates for implicitly supporting the heatmap and self-consistency branches towards fine-grained deepfake detection. Although some attempts have been made to exploit multi-scale features~\cite{caddm}, no previous works have considered FPN in the context of deepfake detection. 

Over the last years, several FPN variants have been proposed for numerous computer vision tasks~\cite{focal_loss, fcos, fpn_seg, fpn_obdet}. Nevertheless, these FPN-based methods usually lead to the generation of redundant features, which might, in turn, lead to the overfitting of the model~\cite{ayinde2019regularizing}. Moreover, as described in Section~\ref{sec:intro}, small discrepancies are gradually eliminated through the successive convolution blocks~\cite{multi-attentional}, going from high-resolution low-level to low-resolution high-level features. Consequently, the last block outputs usually contain global features where local artifact-sensitive features might be discarded. To overcome this issue, we introduce a new alternative referred to as Enhanced Feature Pyramid Network (E-FPN) that is integrated in the proposed LAA-Net architecture. The E-FPN goal is to propagate relevant information from high to low-resolution feature representations.

As shown in Figure~\ref{fig:efpn}, we denote the output shape of the $N-1$ latest layers by $(n^{(l)}, D^{(l)}, D^{(l)})$  with $l$ $\in$ $ [\![ 2,N]\!]$. For the sake of simplicity, we assume that the shape of the feature maps is square. For a given layer $l$, $n^{(l)}, D^{(l)}$ and $\mathbf F^{(l)}$ correspond, respectively, to its feature dimension, its height and width, and its output features. For strengthening the textural information in the ultimate layer $\mathbf F^{(N)}$, we propose to take advantage of the features generated by previous layers $\mathbf F^{(l)}$ with $l~\in$~$ [\![ 2, N-1]\!]$. Concretely, for each layer $l$, a convolution followed by a transpose convolution is applied to $\mathbf F^{(l+1)}$. The obtained features are denoted by $\mathbf E^{(l)}$ and have the same shape as $\mathbf F^{(l)}$. Then, a sigmoid function is applied to $\mathbf E^{(l)}$ returning probabilities. The latter indicates the pixels that contributed to the final decision. For enriching $\mathbf F^{(l+1)}$ while avoiding redundancy related to the most contributing pixels, the features $\mathbf F^{(l)}$ are filtered by computing $(1 - \text{sigmoid}(\mathbf E^{(l)}))^{\gamma_w}$ resulting in a weighted mask. The latter is concatenated along the same axis with $\mathbf E^{(l)}$ for obtaining the final features.
This operation is iterated for all the layers $l\in$ $[\![ 2, N-1]\!]$. In summary, the final representation $\mathbf  F'^{(l)}$  is obtained as follows,
\begin{equation}
   \mathbf  F'^{(l)}= (\mathbf F^{(l)} \odot  (1 - \text{sigmoid}(\mathbf  E^{(l)}))^{\gamma_w}\oplus \mathbf  E^{(l)}) \ ,
\end{equation}
where $\mathbf  E^{(l)}=\mathfrak{T}(f(\mathbf F'^{(l+1)})$ with $\mathbf F'^{(l+1)} = \mathbf F^{(l+1)}$ if $l = N-1$, such that $f$ and $\mathfrak{T}$,  are respectively the convolution and transpose convolution operators, and $\oplus$ refer to the concatenation operator. The hyper-parameter $\gamma_w$ is set to 1 in all our experiments. The relevance of E-FPN in the context of deepfake detection is experimentally demonstrated in Section~\ref{sec:experiment}, as compared to the traditional FPN.

%% file: sec/4_experiment.tex
\begin{table*}[]
\centering
\resizebox{\textwidth}{!}{
\begin{tabular}{|c|cc||c||HHHHccHH|ccHH|ccHH|cc|}
\hline
\multirow{4}{*}{Method} &\multicolumn{2}{c|}{Training set} & \multicolumn{19}{c|}{Test set}\\
\cline{2-22}
& \multirow{3}{*}{Real} & \multirow{3}{*}{Fake} & In-dataset & \multicolumn{18}{c|}{Cross-dataset}\\
\cline{4-22}
& & &FF++ & \multicolumn{4}{H}{CDF1} & \multicolumn{4}{c|}{CDF2} & \multicolumn{4}{c|}{DFW} & \multicolumn{4}{c|}{DFD} &\multicolumn{2}{c|}{DFDC}\\

 & &  & AUC (\%) & AUC (\%) & AP (\%) & AR  & mF1  & AUC (\%) & AP (\%) & AR  & mF1 & AUC (\%) & AP (\%) & AR & mF1 & AUC (\%) & AP (\%) & AR & mF1 & AUC (\%) & AP (\%)\\
\hline
\hline
Xception~\cite{ff++} & $\checkmark$ & $\checkmark$ & 99.09 & 58.81 & 65.59 & 55.58 & {60.17} & 61.18 & 66.93 & 52.40 & {58.78} & 65.29 & 55.37 & 57.99 & {56.65} & 89.75 & 85.48 & 79.34 & 82.29 & 69.90 & 91.98\\

FaceXRay+BI~\cite{fxray} & $\checkmark$ & $\checkmark$ & 99.20 & 80.58 & 73.33 & - & - & 79.5 & - & - & {-} & - & - & - & {-} & 95.40 & 93.34 & - & - & 65.5 & - \\

LRNet~\cite{lrnet} & $\checkmark$ & $\checkmark$ & - & 52.84 & - & - & {-} & 53.20 & - & - & - & - & - & - & - & 52.29 & - & - & - & - & -\\

LocalRL~\cite{localRL} & $\checkmark$ & $\checkmark$ & 99.92 & - & - & - & {-} & 78.26 & - & - & {-} & - & - & - & {-} & 89.24 & - & - & - &76.53 & -\\

TI$^2$Net~\cite{ti2net} & $\checkmark$ & $\checkmark$ & - & 66.65 & - & - & {-} & 68.22 & - & - & {-} & - & - & - & {-} & 72.03 & - & - & - & - & -\\

Multi-attentional~\cite{multi-attentional} & $\checkmark$ & $\checkmark$ & - & 69.14 & 74.03 & 52.70 & {61.57} & 68.26 & 75.25 & 52.40 & {61.78} & 73.56 & 73.79 & 63.38 & {68.19} & 92.95 & 96.51 & 60.76 & 74.57 & 63.02 & - \\

RECCE~\cite{ete_recons} & $\checkmark$ & $\checkmark$ & - & 49.96 & 63.04 & 50.87 & {56.31} & 70.93 & 70.35 & 59.48 & {64.46} & 68.16 & 54.41 & 56.59 & {55.48} & 98.26 & 79.42 & 69.57 & 74.17 & - & -\\

SFDG~\cite{sfdg} & $\checkmark$ & $\checkmark$ & 99.53 & - & - & - & {-} & 75.83 & - & - & {-} & 69.27 & - & - & {-} & 88.00 & - & - & - & 73.63 & -\\

EIC+IIE~\cite{eic_iie} & $\checkmark$ & $\checkmark$ & 99.32 & - & - & - & {-} & 83.80 & - & - & {-} & - & - & - & {-} & 93.92 & - & - & - & 81.23 & -\\

AltFreezing~\cite{altfreezing} & $\checkmark$ & $\checkmark$ & 98.6 & - & - & - & - & 89.50 & - & - & {-} & - & - & - & {-} & 98.50 & - & - & - & - & -\\

CADDM~\cite{caddm} & $\checkmark$ & $\checkmark$ & 99.79 & 89.36 & \underline{93.25} & \underline{81.41} & {\underline{86.93}} & \underline{93.88} & 91.12 & 77.00 & {83.46} & \underline{74.48} & \underline{75.23} & \underline{65.26} & {\underline{69.89}} & 99.03 & \underline{99.59} & 82.17 & 90.04 & - & -\\

UCF~\cite{ucf} & $\checkmark$ & $\checkmark$ & - & - & - & - & - & 82.4 & - & - & - & - & - & - & - & 94.5 & - & - & - & 80.5 & - \\

Controllable GS~\cite{cgs} & $\checkmark$ & $\checkmark$ & - & - & - & - & - & 84.97 & - & - & - & - & - & - & - & - & - & - & - & 81.65 & - \\

\hline

PCL+I2G~\cite{cstency_learning} & $\checkmark$ & & 99.11 & \textbf{98.30} & - & - & {-} & 90.03 & - & - & {-} & - & - & - & {-} & 99.07 & - & - & - & 74.27 & -\\

SBI~\cite{sbi} & $\checkmark$ & & 99.64 & 92.53 & 79.91 & 79.16 & {79.53} & 93.18 & 85.16 & \underline{82.68} & {\underline{83.90}} & 67.47 & 55.87 & 55.82 & {55.85} & 97.56 & 92.79 & \underline{89.49} & 91.11 & 86.15 & 93.24 \\

AUNet~\cite{aunet} & $\checkmark$ & & 99.46 & - & - & - & {-} & 92.77 & - & - & {-} & - & - & - & {-} & \underline{99.22} & - & - & - & \underline{86.16} & -\\

\hline
\hline
Ours (w/ BI) & $\checkmark$ & & \underline{99.95} & 92.46 & 95.54 & 50.0 & 65.64 & 86.28 & \underline{91.93} & 50.0 & 64.77 & 57.13 & 56.89 & 50.12 & {{53.29}} & \textbf{99.51} & \textbf{99.80} & \textbf{95.47} & \textbf{97.59} & 69.69 & \underline{93.67} \\ 
\hdashline

Ours (w/ SBI) & $\checkmark$ &  & \textbf{99.96} & \underline{93.11} & \textbf{95.64} & \textbf{89.78} & {\textbf{92.62}} & \textbf{95.40} & \textbf{97.64} & \textbf{87.71} & {\textbf{92.41}} & \textbf{80.03} & \textbf{81.08} & \textbf{65.66} & {\textbf{72.56}} & 98.43 & 99.40 & 88.55 & \underline{93.64} & \textbf{86.94} & \textbf{97.70} \\
\hline
\end{tabular}%
}
\vspace{-1mm}
\caption{In-dataset and Cross-dataset evaluation in terms of AUC and AP on multiple deepfake datasets. \textbf{Bold} and \underline{Underlined} highlight the best and the second-best performance, respectively.
}
\vspace{-3mm}
\label{tabl:cross_auc_full_metrics_2}
\end{table*}

\section{Experiments}
In this section, we start by presenting the experimental settings. Then, we compare the performance of LAA-Net to SoA methods, both qualitatively and quantitatively. Finally, we conduct an ablation study to validate the different components of LAA-Net.

\label{sec:experiment}
\subsection{Experimental Settings}
\label{subsec:data_eval_metric}
\noindent\textbf{Datasets.}\label{subsec:datasets}
The FF++~\cite{ff++} dataset is used for training and validation. In our experiments, we follow the standard splitting protocol of~\cite{ff++}. This dataset contains $1000$ original videos and $4000$ fake videos generated by four different manipulation methods, namely, Deepfakes (DF)~\cite{deepfake}, Face2Face (F2F)~\cite{face2face}, FaceSwap (FS)~\cite{faceswap}, and NeuralTextures (NT)~\cite{neutex}. In the training process, we utilize real images only to dynamically generate pseudo-fakes, as discussed in Section~\ref{sec:method}. To evaluate the generalization capability of the proposed approach as well as its robustness to high-quality deepfakes, we test the trained model on four datasets incorporating different quality of deepfakes, namely, Celeb-DFv2~\cite{celeb_df} (CDF2), DeepFake Detection~\cite{dfd} (DFD), DeepFake Detection Challenge~\cite{dfdcp} (DFDC) and Wild Deepfake~\cite{wdf} (DFW). To assess the quality of the considered datasets, we compute the Mask-SSIM\footref{mssim_defi} for each benchmark. In particular, CDF2~\cite{celeb_df} is formed by the most realistic deepfakes with an average Mask-SSIM~\cite{mssim_pose} value of $0.92$, followed by DFD and DFDC with an average Mask-SSIM of $0.88$ and $0.84$, respectively. We note that computing the Mask-SSIM~\cite{celeb_df} for DFW was not possible since real and fake images are not paired.

\vspace{1mm}
\noindent\textbf{Evaluation Metrics.}
\label{subsec:eval_metrics}
To compare the performance of LAA-Net with the state-of-the-art, we report the common Area Under the Curve (AUC) metric at the video-level and the Average Precision (AP) as in~\cite{fxray, cstency_learning, sbi, caddm}. More metrics, namely, Average Recall (AR) and mean F1-score (mF1) are provided in supplementary materials.

\vspace{1mm}
\noindent\textbf{Implementation Details.}\label{subsec:preprocessing}
To train our model, $128$ training and $32$ validation frames are used. 
RetinaNet~\cite{focal_loss} is used to crop faces with a conservative enlargement (by a factor of $1.25$) around the face center. Note that all the cropped images are then resized to $384 \times 384$. In addition, $68$ facial landmarks are extracted per frame using Dlib~\cite{dlib}. We adopt the EFNB4 variant of the EfficientNet~\cite{efn_net} pretrained on ImageNet~\cite{imagenet}.
For each training epoch, $8$ frames are dynamically selected and used for online pseudo-fake generation.
The model is trained for $100$ epochs with the SAM optimizer~\cite{sam}, a weight decay of $10^{-4}$, and a batch size of $16$. We apply a learning rate scheduler that increases from $5. 10^{-5}$ to $2. 10^{-4}$ in the first quarter of the training and then decays to zero in the remaining quarters. We freeze the backbone at the first $6$ epochs and only train the remaining layers. For data augmentation, we apply horizontal flipping, random cropping, random scaling, random erasing~\cite{random_erasing}, color jittering, Gaussian noise, blurring, and JPEG compression. The parameters $\lambda_1$ and $\lambda_2$, defined in Eq.~\eqref{equa:total_loss}, are set to $10$ and $100$. Furthermore, label smoothing~\cite{label_smoothing} is utilized as a regularizer. To generate pseudo-fakes, two blending synthesis techniques are considered, namely, Blended Images (BI)~\cite{fxray} and Self-Blended Images (SBI)~\cite{sbi}. All experiments are carried out using a GPU Tesla V-$100$.

\subsection{Comparison with State-of-the-art }
\noindent\textbf{In-dataset Evaluation.} 
We compare the performance of LAA-Net to existing methods under the in-dataset protocol of ~\cite{cstency_learning, caddm, sbi, aunet, sfdg, altfreezing}. The first column in Table~\ref{tabl:cross_auc_full_metrics_2} reports the obtained results on the testing set of FF++. It can be seen that all methods achieve competitive performance on the forgeries of the FF++ dataset. Our method combined with SBI outperforms all methods with an AUC of $99.96$\%, while using only real data for training.

\begin{table}[!t]
\centering
\resizebox{\columnwidth}{!}{%
\begin{tabular}{|c|c|cccccc|}
\hline
{Method} & {Fake} & Saturation & Contrast & Block & Noise & Blur & Pixel \\ 
\hline
\hline
{Xception~\cite{xception}} & {$\checkmark$} & 99.3 & 98.6 & 99.7 & 53.8 & 60.2 & 74.2 \\

{FaceXray~\cite{fxray}} & {$\checkmark$} & 97.6 & 88.5 & 99.1 & 49.8 & 63.8 & 88.6 \\

{LipForensics~\cite{lipforensics}} & {$\checkmark$} & \underline{99.9} & 99.6 & 87.4 & \underline{73.8} & 96.1 & 95.6 \\

{CADDM~\cite{caddm}} & {$\checkmark$} & 99.6 & \underline{99.8} & \underline{99.8} & \textbf{87.4} & \textbf{99.0} & \underline{98.8} \\
\hline
\hline

{Ours} & {} & \textbf{99.96} & \textbf{99.96} & \textbf{99.96} & 53.9 & \underline{98.22} & \textbf{99.80} \\ 
\hline
\end{tabular}%
}
\caption{Robustness inspection on the FF++ with different types of perturbation. \textbf{Bold} and \underline{Underline} highlight the best and the second-best performance, respectively.}
\label{tabl:ff_noise_auc}
\end{table}

\begin{table}[]
\resizebox{\columnwidth}{!}{
\begin{tabular}{|ccc|Hcccc|c|}
\hline
 \multirow{2}{*}{C} & \multirow{2}{*}{H} & \multirow{2}{*}{E-FPN} & \multicolumn{6}{c|}{Test set AUC (\%)} \\ 
 \cline{4-9} 
 &  &  & {CDF1} & CDF2 & DFD & DFDC & \multicolumn{1}{c|}{DFW} & Avg. \\ 
\hline
\hline
$\times$ & $\times$ & $\times$ & - & 74.54 & 92.24 & 70.81 & 59.81 & 74.35 \\

$\times$ & $\checkmark$ & $\times$ & - & 80.89 & 94.53 & 77.93 & 67.12 & 80.12($\uparrow$5.77) \\

$\times$ & $\times$ & $\checkmark$ & - & 84.21 & 95.03 & 80.68 & 65.47 & 81.35($\uparrow$7.00) \\

$\times$ & $\checkmark$ & $\checkmark$ & \textbf{97.82} & \textbf{95.56} & \textbf{98.54} & \underline{82.21} & \multicolumn{1}{c|}{\underline{74.98}} & \underline{87.82}($\uparrow$13.47) \\

$\checkmark$ & $\times$ & $\checkmark$ & 84.13 & 79.87 & 94.60 & 71.70 & \multicolumn{1}{c|}{72.47} & 79.66($\uparrow$5.31) \\

$\checkmark$ & $\checkmark$ & $\times$ & 92.72 & 91.56 & 98.27 & 78.35 & \multicolumn{1}{c|}{73.02} & 85.30($\uparrow$10.95) \\

$\checkmark$ & $\checkmark$ & $\checkmark$ & \underline{93.11} & \underline{95.40} & \underline{98.43} & \textbf{86.94} & \multicolumn{1}{c|}{\textbf{80.03}} & \textbf{90.20($\uparrow$15.85)} \\ 
\hline
\end{tabular}%
}
\caption{Ablation study under the cross-dataset setup of the Consistency branch (C), Heatmap branch (H), and E-FPN.}
\label{tabl:laa_components_eval}
\end{table}

\begin{table*}[ht]
\centering
\scalebox{0.9}{
\begin{tabular}{|c|c|cccc|HHcc|cc|cc|cc|}
\hline
& \multicolumn{5}{c|}{EFNB4} & \multicolumn{10}{c|}{Test Set AUC (\%)} \\
\hline
\multicolumn{2}{|c|}{} & \multicolumn{4}{c|}{E-FPN Integration} & \multicolumn{2}{H}{CDF1} & \multicolumn{2}{c|}{CDF2} & \multicolumn{2}{c|}{DFD} & \multicolumn{2}{c|}{DFW} & \multicolumn{2}{c|}{DFDC} \\
\hline
& {$\mathbf F^{(6)}$} &$\mathbf F^{(5)}$ & $\mathbf F^{(4)}$ & $\mathbf F^{(3)}$ & {$\mathbf F^{(2)}$} & FPN & {E-FPN} & FPN & {E-FPN} & FPN & {E-FPN} & FPN & {E-FPN} & FPN & E-FPN \\
\hline
\hline
(a) & {$\surd$} & & & & {} & \multicolumn{2}{H}{92.72} & \multicolumn{2}{c|}{91.56} & \multicolumn{2}{c|}{98.27} & \multicolumn{2}{c|}{73.02} & \multicolumn{2}{c|}{78.35} \\

(b) & {$\surd$} & $\surd$ & & & {} & 93.37 & {{93.81}} & {93.42} & {91.79} & {98.59} & {97.12} & {73.78} & {71.39} & {78.40} & 75.80 \\

(c) & {$\surd$} & $\surd$ & $\surd$ & & {} & 89.45 & {\underline{{95.34}}} & 88.72 & {{92.86}} & 97.96 & {\textbf{{98.95}}} & 69.40 & {\underline{{74.93}}} & 71.91 & \underline{83.97} \\

(d) & {$\surd$} & $\surd$ & $\surd$ & $\surd$ & {} & 88.97 & {{93.11}} & 88.35 & {\textbf{{95.40}}} & \underline{98.89} & {98.43} & 70.94 & {\textbf{{80.03}}} & 79.02 & \textbf{{86.94}} \\

(e) & {$\surd$} & $\surd$ & $\surd$ & $\surd$ & {$\surd$} & 92.59 & {\textbf{{95.64}}} & 92.16 & {\underline{{94.22}}} & 96.58 & {97.31}& 65.17 & {{72.54}} & 74.31 & {82.90} \\

\hline
\hline

\multicolumn{6}{|c|}{Avg} &  & {\textbf{{-}}} & 90.84 & \textit{93.16} & \textit{98.06} & 98.02 & 70.46 & \textit{74.38} & 76.40 & \textit{81.59} \\

\hline
\end{tabular}
}
\vspace{-2mm}
\caption{Traditional FPN versus E-FPN, using the SBI-based data synthesis under the cross-dataset protocol.  \textbf{Bold} and \underline{Underline} indicate the best and the second-best performance, respectively. We report the results when integrating features $\mathbf F^{(i)}$ from different layers.  }
\vspace{-3mm}
\label{tabl:efpn_tfpn}
\end{table*}

\vspace{1.5mm}
\noindent\textbf{Cross-dataset Evaluation.}
We evaluate LAA-Net under the challenging cross-dataset setup ~\cite{ete_recons, sfdg}. Table~\ref{tabl:cross_auc_full_metrics_2} reports the obtained results on CDF2, DFW, DFD, and DFDC, respectively. 
It can be noted that LAA-Net achieves state-of-the-art results on the four considered benchmarks, thereby demonstrating its robustness to different quality of deepfakes. The best performance is reached using SBI as a data synthesis, confirming the importance of modeling subtle artifacts. The performance of LAA-Net (w/BI) is slightly superior to LAA-Net (w/SBI) only on DFD, with an improvement of $1.08$\%  and $0.4$\% of AUC and AP, respectively. A plausible explanation could be the fact that deepfake detection in DFD is less challenging. In fact, numerous methods report AUC and AP scores exceeding $98$\%. 

Furthermore, LAA-Net clearly outperforms attention-based approaches such as Multi-attentional~\cite{multi-attentional} and SFDG~\cite{sfdg} by a margin of $27.14$\% and $19.57$\% in terms of AUC and AP on CDF2, respectively. This confirms the superior generalization capabilities of LAA-Net as compared to~\cite{multi-attentional,sfdg}. These results are further supported by high AR and mF1, which are provided in the supplementary materials.

\vspace{1.5mm}
\noindent\textbf{Robustness to Perturbations.}\label{noise_robust} Since deepfake videos are easily altered on various social platforms, the robustness of LAA-Net against some common perturbations is investigated. Following the same settings of~\cite{lipforensics, caddm}, we evaluate the performance of LAA-Net on FF++~\cite{ff++} by applying different corruptions. The results are reported in Table~\ref{tabl:ff_noise_auc}. 
As our method focuses on vulnerable points, it can be seen that color-related changes such as saturation and contrast do not impact the performance. 
However, the proposed method is extremely sensitive to structural perturbations such as Gaussian Noise. In future work, strategies for ensuring more robustness to structural perturbations will be investigated. For instance, denoising methods~\cite{sanet, zs_n2n} will be considered for solving this issue.

\begin{figure}[t]
    \centering
    \includegraphics[width=0.80\linewidth]{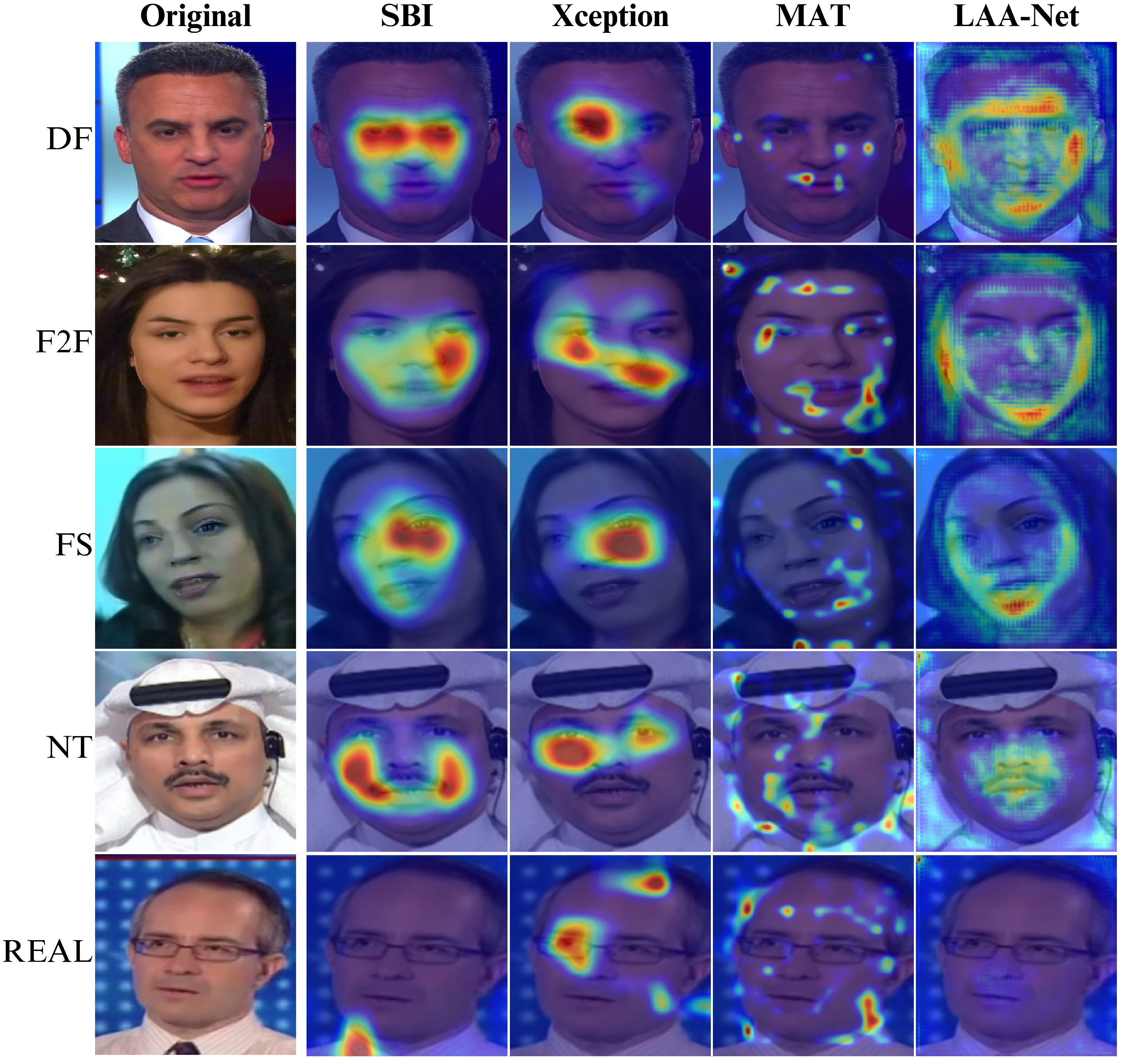}
    \vspace{-2mm}
    \caption{Grad-CAM~\cite{gradCAM} visualization on different types of manipulation from FF++~\cite{ff++}. LAA-Net is compared to SBI~\cite{sbi}, Xception~\cite{ff++}, and MAT~\cite{multi-attentional}.}
    \vspace{-3mm}
\label{fig:qualitative_results}
\end{figure}

\vspace{1.5mm}
\noindent\textbf{Qualitative Results.}\label{abl:qua_res}
We provide Grad-CAMs~\cite{gradCAM} in Figure~\ref{fig:qualitative_results}, to visualize the image regions in deepfakes that are activated by LAA-Net, SBI~\cite{sbi}, Xception~\cite{ff++}, and Multi-attentional (MAT)~\cite{multi-attentional} on FF++~\cite{ff++}. Generally, attention-based methods such as MAT~\cite{multi-attentional} and LAA-Net focus more on localized regions. However, in some cases, MAT~\cite{multi-attentional} concentrates on irrelevant regions such as the background or the inner face areas. Conversely, LAA-Net consistently identifies blending artifacts and shows interesting capabilities on mouth-rendered Neural Textures (NT).

%% file: sec/4_x_ablation.tex
\subsection{Ablation Study}
\begin{figure}[t]
    \centering
    \includegraphics[width=\linewidth]{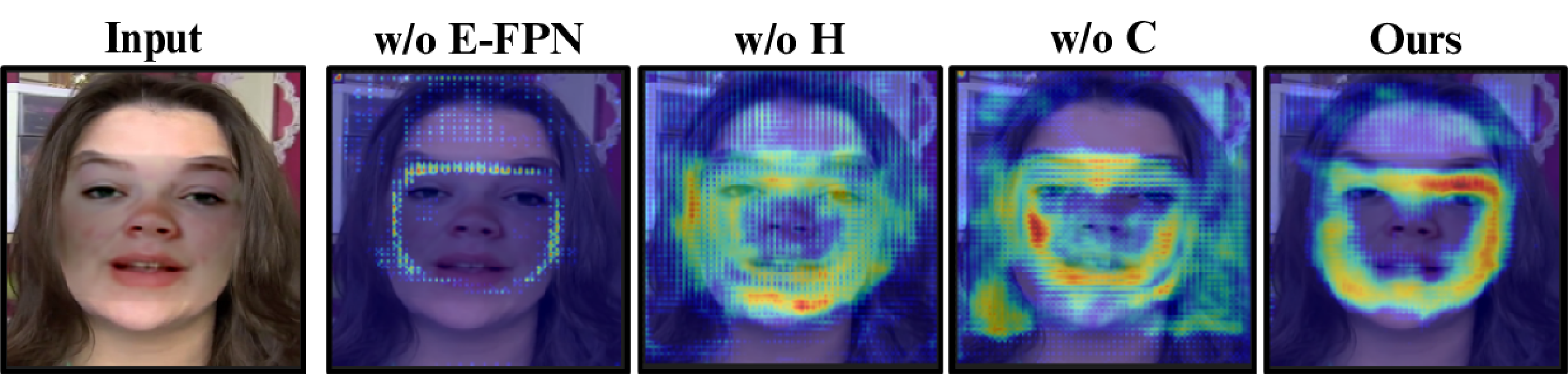}
    \vspace{-2mm}
    \caption{GradCAM~\cite{gradCAM} visualization of different components in LAA-Net. w/o E-FPN, w/o H, and w/o C refer to ablating E-FPN, heatmap branch, and self-consistency branch, respectively.}
    \vspace{-3mm}
    \label{fig:efpn_com_abl}
\end{figure}

Table~\ref{tabl:laa_components_eval} reports the cross-dataset performance of LAA-Net when discarding the following components:  E-FPN, the consistency branch denoted by C and the heatmap branch denoted by H. The best performance is reached when all the components are integrated. It can be seen that the proposed explicit attention mechanism through the heatmap branch contributes more to improving the result. A qualitative example visualizing Grad-CAMs~\cite{gradCAM} with different components of LAA-Net is also given in Figure~\ref{fig:efpn_com_abl}. The illustration clearly shows that by combining the three components, the network activates more precisely the blending region.

\subsection{ E-FPN versus Traditional FPN} 
To assess the effectiveness of the low-level features injected by E-FPN into the final feature representation, we combine different feature levels and compare the results of E-FPN and traditional FPN~\cite{fpn_obdet, focal_loss} in Table~\ref{tabl:efpn_tfpn}. It can be seen that in general E-FPN outperforms FPN except for $\mathbf{F}^{(5)}$. This confirms the relevance of employing multi-scale features and the need for reducing their redundancy in the context of deepfake detection.

\subsection{Sensitivity Analysis}
In this subsection, we analyze the impact of the two hyperparameters, $\lambda_1$ and $\lambda_2$ given in Eq.~\eqref{equa:total_loss}. Table~\ref{tabl:blance_factor_eval} shows the experimental results for different values of $\lambda_1$ and $\lambda_2$. It can be noted that our model is robust to different hyperparameter values, with the best average performance obtained with $\lambda_1=10$ and $\lambda_2=100$. 
\begin{table}[ht]
\centering
\scalebox{0.87}{
\begin{tabular}{|cc|Hccc|c|} 
\hline
 \multirow{2}{*}{$\lambda_1$} & \multirow{2}{*}{$\lambda_2$} & \multicolumn{5}{c|}{Test Set AUC (\%)} \\
 \cline{3-7}
                                                      & & {CDF1} & CDF2 & DFDC & {DFW} & Avg           \\ 
\hline
\hline
1 & 1 & {94.20} & 90.69 & 78.12 & {70.98} & 79.93 \\
10 & 10 & {\underline{96.60}} & \textbf{95.73} & \underline{85.87} & 73.56 & \underline{85.05} \\
100 & 100 & {\textbf{97.30}} & 93.72 & 78.60 & {75.25} & 82.52 \\
100 & 10 & {93.29} & 93.05 & 83.86 & {\underline{76.72}} & 84.54 \\
10 & 100 & {93.11} & \underline{95.40} & \textbf{86.94} & {\textbf{80.03}} & \textbf{87.46} \\
\hline
\end{tabular}
}
\vspace{-2mm}
\caption{Sensitivity analysis: The impact of the hyper-parameters $\lambda_1$ and $\lambda_2$ using the cross-dataset protocol on three datasets in terms of AUC.}
\vspace{-5mm}
\label{tabl:blance_factor_eval}
\end{table}

%% file: sec/6_conclusion.tex
\vspace{-1mm}
\section{Conclusion}
\label{sec:conclu}

In this paper, a fine-grained deepfake detection method called LAA-Net is introduced with the aim of detecting high-quality deepfakes while remaining generic to unseen manipulations. For that purpose, two different components are proposed. On the one hand, we argue that by making the network focus on the most vulnerable points, we can detect both global and subtle artifacts. To this end, an explicit attention mechanism within a multi-task learning framework is used. In addition to the binary classification branch, heatmap and self-consistency branches are defined with respect to the vulnerable points. On the other hand, a novel E-FPN module for aggregating multi-scale features is proposed; hence enabling the integration of more localized features. The results reported on several benchmarks show the superiority of LAA-Net as compared to the state-of-the-art, including attention-based methods. In future works, strategies for improving the robustness to noise will be investigated. In addition, an attempt to extend this idea by taking into account the temporal dimension will be explored.

\vspace{3mm}
\noindent\textbf{\large{Acknowledgment}}
\label{sec:acknowledge}
\vspace{1mm}

This work is supported by the Luxembourg National Research Fund, under the BRIDGES2021/IS/16353350/FaKeDeTeR and UNFAKE, ref.16763798 projects, and by POST Luxembourg.

%% file: sec/X_suppl.tex

\section*{\centering{Overview}}
This document provides supplementary material complementing the main manuscript. It is structured as follows. First, the computation of the self-consistency loss and the ground truth generation of heatmaps are described. Second, more quantitative and qualitative results are provided. In particular, additional metrics are reported for both in-dataset and cross-dataset settings. Moreover, qualitative results comparing E-FPN and FPN are shown.

\section{Self-Consistency Loss}

\begin{figure}[!h]
    \centering
    \includegraphics[width=\linewidth]{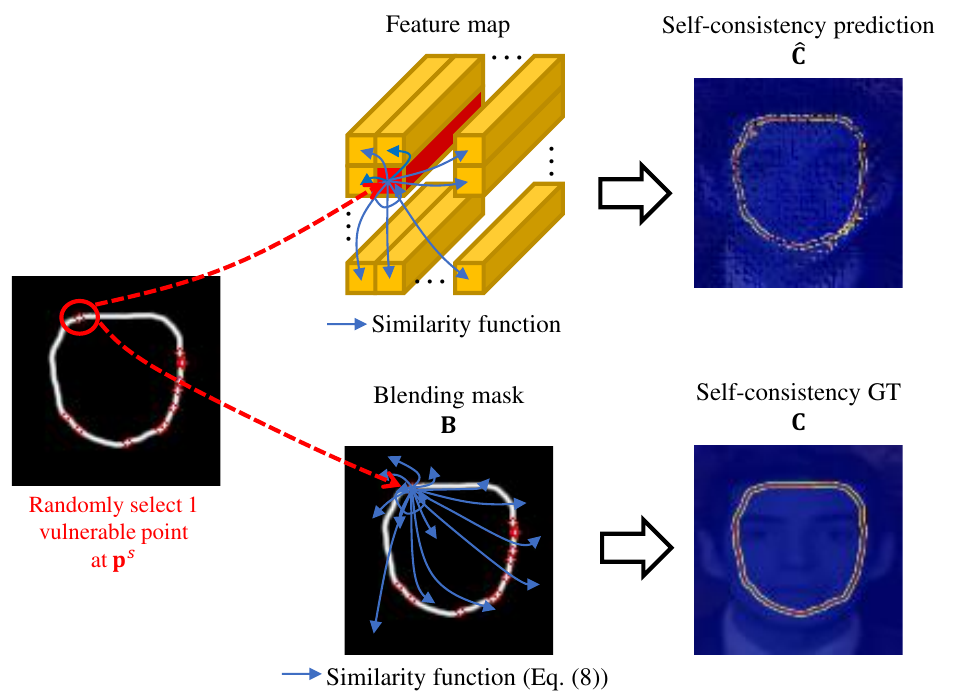}
    \caption{In order to generate the consistency map prediction $\hat{\mathbf{C}}$ as well as the associated ground truth $\mathbf{C}$, we first randomly select a vulnerable point located at $\mathbf{p}^s$. For computing $\hat{\mathbf{C}}$, we measure the similarity between the feature at $\mathbf{p}^s$ (red block) and the features generated from every point. Namely, we use the similarity function in \cite{cstency_learning}. As for $\mathbf{C}$, we measure the consistency values between the pixel at the $\mathbf{p}^s$ and all pixels in $\mathbf{B}$, as also described in Eq. (\textcolor{red}{7}) of the manuscript.}
    \label{fig:consistency}
\end{figure}

\begin{figure}[!h]
    \centering
    \includegraphics[width=\linewidth]{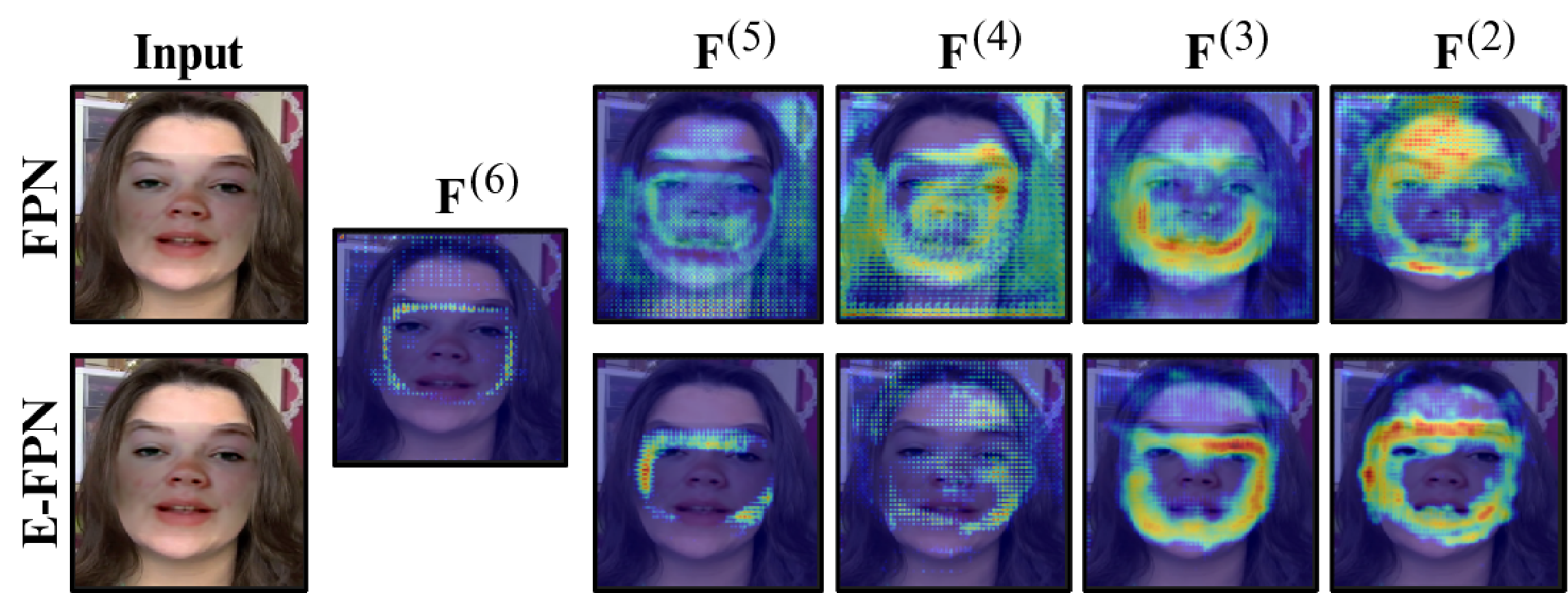}
    \caption{Feature visualization by gradCAM~\cite{gradCAM} between \textit{E-FPN} and \textit{FPN} with different integration of multi-scale layers. It shows that E-FPN can focus better on artifacts as compared to FPN. The setup details are provided in Table \textcolor{red}{4} as shown in the manuscript.}
    \label{fig:efpn_fpn_abl}
\end{figure}

To clarify the calculation of the self-consistency loss, we show Figure \ref{fig:consistency}, which illustrates the generation process of the predicted and the ground-truth, $\hat{\mathbf{C}}$ and $\mathbf{C}$, respectively.  The self-consistency loss is a binary cross entropy loss between $\hat{\mathbf{C}}$ and $\mathbf{C}$.

\begin{table}[]
\centering
\resizebox{\columnwidth}{!}{%
\begin{tabular}{ccccccccc}
\hline
\multirow{2}{*}{Method} & \multicolumn{2}{|c|}{Training Set} & \multicolumn{5}{c}{FF++~\cite{ff++}} \\
\cline{2-8}
 & \multicolumn{1}{|c}{Real} & \multicolumn{1}{c|}{Fake} & ACC & AUC & AP & AR & mF1 \\ 
\hline
\hline
Ours w/ BI~\cite{fxray} & \multicolumn{1}{|c}{$\checkmark$} & \multicolumn{1}{c|}{} & 99.03 & 99.95 & 99.99 & 99.21 & 99.60 \\
Ours w/ SBI~\cite{sbi} & \multicolumn{1}{|c}{$\checkmark$} & \multicolumn{1}{c|}{} & 99.04 & 99.96 & 99.99 & 99.29 & 99.64 \\
\hline
\end{tabular}%
}
\caption{In-dataset evaluation on FF++~\cite{ff++} reported by ACC, AUC, AP, AR, and mF1.}
\label{tabl:ff_full_metrics_suppl}
\end{table}

\section{Ground Truth Generation of Heatmaps}

\begin{figure*}[!h]
    \centering
    \includegraphics[width=\linewidth]{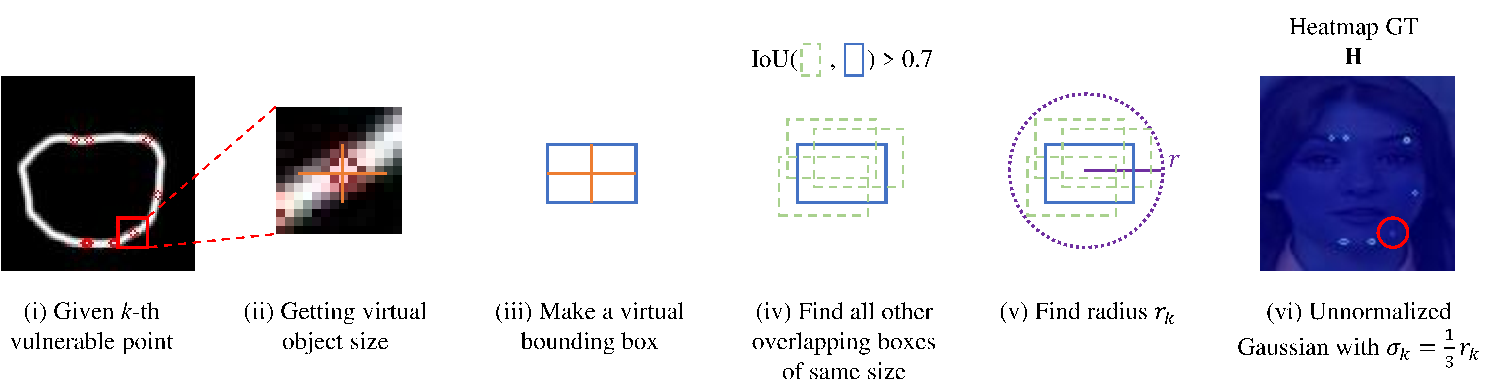}
    \caption{
    The generation process of ground truth heatmaps by producing using an \textit{Unnormalized Gaussian Distribution} given a selected vulnerable point.
    }
    \label{fig:heatmap_gt_generation}
\end{figure*}

In this section, we provide more details regarding the generation of ground-truth heatmaps, described in Section \textcolor{red}{3.1.2}. Firstly, a $k$-th vulnerable point, denoted as $\mathbf{p}^k$, is selected, as shown in Figure \ref{fig:heatmap_gt_generation} (i). Secondly, we measure the height and the width of the blending mask $\mathbf{B}$ at the point $\mathbf{p}^k$ shown as orange lines in Figure \ref{fig:heatmap_gt_generation} (ii). Using the calculated distances, a virtual bounding box is created, indicated by the blue box in Figure \ref{fig:heatmap_gt_generation} (iii). Then, we identify overlapping boxes, illustrated by dashed-line green boxes in Figure \ref{fig:heatmap_gt_generation} (iv), with the Intersection over Union (IoU) greater than a threshold ($t=0.7)$ compared to the virtual bounding box. A radius $r_k$ (solid purple line in Figure \ref{fig:heatmap_gt_generation} (v)) is calculated by forming a tight circle encompassing all these boxes. Finally, an \textit{Unnormalized Gaussian Distribution}, 
shown as a red circle in Figure \ref{fig:heatmap_gt_generation} (vi), is generated with a standard deviation $\sigma_k = \frac{1}{3}r_k$ (Eq. (\textcolor{red}{4}) of the manuscript).
The steps are repeated for every vulnerable point $k \in [\![ 1, \text{card}(\mathcal P)]\!]$. The final $\mathbf{H}$ is the superimposition of all  $g_{ij}^k$.

\begin{table*}[]
\centering
\resizebox{\textwidth}{!}{
\begin{tabular}{|c| c| HHHHcccccccccccccccc|}
\hline
\multirow{3}{*}{Method} & \multirow{3}{*}{Fake} & \multicolumn{20}{c|}{Test set (\%)} \\
\cline{3-22}
 &  & \multicolumn{4}{H}{CDF1} & \multicolumn{4}{c|}{CDF2} & \multicolumn{4}{c|}{DFW} & \multicolumn{4}{c|}{DFD} & \multicolumn{4}{c|}{DFDC} \\
 
\cline{3-22}
 &  & AUC & AP & AR & mF1 & AUC & AP & AR & \multicolumn{1}{c|}{mF1} & AUC & AP & AR & \multicolumn{1}{c|}{mF1} & AUC & AP & AR & \multicolumn{1}{c|}{mF1} & AUC & AP & AR & \multicolumn{1}{c|}{mF1} \\
 
\hline
\hline
Xception~\cite{ff++} & $\checkmark$ & 58.81 & 65.59 & 55.58 & 60.17 & 61.18 & 66.93 & 52.40 & \multicolumn{1}{c|}{58.78} & 65.29 & 55.37 & 57.99 & \multicolumn{1}{c|}{56.65} & 89.75 & 85.48 & 79.34 & \multicolumn{1}{c|}{82.29} & 69.90 & 91.98 & 67.07 & 77.57 \\

FaceXRay+BI~\cite{fxray} & $\checkmark$ & 80.58 & 73.33 & - & - & 79.5 & - & - & \multicolumn{1}{c|}{-} & - & - & - & \multicolumn{1}{c|}{-} & 95.40 & 93.34 & - & \multicolumn{1}{c|}{-} & 65.5 & - & - & - \\

LRNet~\cite{lrnet} & $\checkmark$ & 52.84 & - & - & - & 53.20 & - & - & \multicolumn{1}{c|}{-} & - & - & - & \multicolumn{1}{c|}{-} & 52.29 & - & - & \multicolumn{1}{c|}{-} & - & - & - & - \\

LocalRL~\cite{localRL} & $\checkmark$ & - & - & - & - & 78.26 & - & - & \multicolumn{1}{c|}{-} & - & - & - & \multicolumn{1}{c|}{-} & 89.24 & - & - & \multicolumn{1}{c|}{-} & 76.53 & - & - & - \\

TI$^2$Net~\cite{ti2net} & $\checkmark$ & 66.65 & - & - & - & 68.22 & - & - & \multicolumn{1}{c|}{-} & - & - & - & \multicolumn{1}{c|}{-} & 72.03 & - & - & \multicolumn{1}{c|}{-} & - & - & - & - \\

Multi-attentional~\cite{multi-attentional} & $\checkmark$ & 69.14 & 74.03 & 52.70 & 61.57 & 68.26 & 75.25 & 52.40 & \multicolumn{1}{c|}{61.78} & 73.56 & 73.79 & 63.38 & \multicolumn{1}{c|}{68.19} & 92.95 & 96.51 & 60.76 & \multicolumn{1}{c|}{74.57} & 63.02 & - & - & - \\

RECCE~\cite{ete_recons} & $\checkmark$ & 49.96 & 63.04 & 50.87 & 56.31 & 70.93 & 70.35 & 59.48 & \multicolumn{1}{c|}{64.46} & 68.16 & 54.41 & 56.59 & \multicolumn{1}{c|}{55.48} & 98.26 & 79.42 & 69.57 & \multicolumn{1}{c|}{74.17} & - & - & - & - \\

SFDG~\cite{sfdg} & $\checkmark$ & - & - & - & - & 75.83 & - & - & \multicolumn{1}{c|}{-} & 69.27 & - & - & \multicolumn{1}{c|}{-} & 88.00 & - & - & \multicolumn{1}{c|}{-} & 73.63 & - & - & - \\

EIC+IIE~\cite{eic_iie} & $\checkmark$ & - & - & - & - & 83.80 & - & - & \multicolumn{1}{c|}{-} & - & - & - & \multicolumn{1}{c|}{-} & 93.92 & - & - & \multicolumn{1}{c|}{-} & 81.23 & - & - & - \\

AltFreezing~\cite{altfreezing} & $\checkmark$ & - & - & - & - & 89.50 & - & - & \multicolumn{1}{c|}{-} & - & - & - & \multicolumn{1}{c|}{-} & 98.50 & - & - & \multicolumn{1}{c|}{-} & - & - & - & - \\

CADDM~\cite{caddm} & $\checkmark$ & 89.36 & \underline{93.25} & \underline{81.41} & 86.93 & \underline{93.88} & 91.12 & 77.00 & \multicolumn{1}{c|}{83.46} & \underline{74.48} & \underline{75.23} & \underline{65.26} & \multicolumn{1}{c|}{\underline{69.89}} & 99.03 & \underline{99.59} & 82.17 & \multicolumn{1}{c|}{90.04} & - & - & - & - \\

UCF~\cite{ucf} & $\checkmark$ & - & - & - & - & 82.4 & - & - & \multicolumn{1}{c|}{-} & - & - & - & \multicolumn{1}{c|}{-} & 94.5 & - & - & \multicolumn{1}{c|}{-} & 80.5 & - & - & - \\

Controllable GS~\cite{cgs} & $\checkmark$ & - & - & - & - & 84.97 & - & - & \multicolumn{1}{c|}{-} & - & - & - & \multicolumn{1}{c|}{-} & - & - & - & \multicolumn{1}{c|}{-} & 81.65 & - & - & - \\

\hline

PCL+I2G~\cite{cstency_learning} &  & \textbf{98.30} & - & - & - & 90.03 & - & - & \multicolumn{1}{c|}{-} & - & - & - & \multicolumn{1}{c|}{-} & 99.07 & - & - & \multicolumn{1}{c|}{-} & 74.27 & - & - & - \\

SBI~\cite{sbi} &  & 92.53 & 79.91 & 79.16 & 79.53 & 93.18 & 85.16 & \underline{82.68} & \multicolumn{1}{c|}{\underline{83.90}} & 67.47 & 55.87 & 55.82 & \multicolumn{1}{c|}{55.85} & 97.56 & 92.79 & \underline{89.49} & \multicolumn{1}{c|}{91.11} & 86.15 & 93.24 & \underline{71.58} & \underline{80.99} \\

AUNet~\cite{aunet} &  & - & - & - & - & 92.77 & - & - & \multicolumn{1}{c|}{-} & - & - & - & \multicolumn{1}{c|}{-} & \underline{99.22} & - & - & \multicolumn{1}{c|}{-} & \underline{86.16} & - & - & - \\

\hline
\hline
Ours (w/ BI) &  & 91.67 & 94.79 & 50.0 & 65.47 & 86.28 & \underline{91.93} & 50.01 & \multicolumn{1}{c|}{64.78} & 57.13 & 56.89 & 50.12 & \multicolumn{1}{c|}{{53.29}} & \textbf{99.51} & \textbf{99.80} & \textbf{95.47} & \multicolumn{1}{c|}{\textbf{97.59}} & 69.69 & \underline{93.67} & 50.12 & 65.30 \\
\hdashline

Ours (w/ SBI) &  & \underline{93.11} & \textbf{95.64} & \textbf{89.78} & 92.62 & \textbf{95.40} & \textbf{97.64} & \textbf{87.71} & \multicolumn{1}{c|}{\textbf{92.41}} & \textbf{80.03} & \textbf{81.08} & \textbf{65.66} & \multicolumn{1}{c|}{\textbf{72.56}} & 98.43 & 99.40 & 88.55 & \multicolumn{1}{c|}{\underline{93.64}} & \textbf{86.94} & \textbf{97.70} & \textbf{73.37} & \textbf{83.81} \\
\hline
\end{tabular}%
}
\caption{Cross-dataset evaluation in terms of AUC, AP, AR, and mF1 (\%) on CDF2~\cite{celeb_df}, DFW~\cite{wdf}, DFD~\cite{dfd}, and DFDC~\cite{dfdcp}. \textbf{Bold} and \underline{underlined} highlight the best and the second-best performance, respectively. \checkmark symbol is used to depict methods that utilized both Real data and Fake data for training.
}
\label{tabl:cross_auc_full_metrics_suppl}
\end{table*}

\section{Additional Results}

In addition to AUC, we provide results using additional metrics, namely, Average Precision (AP), Average Recall (AR), Accuracy (ACC), and mean F1-score (mF1). 

Table~\ref{tabl:ff_full_metrics_suppl} and Table~\ref{tabl:cross_auc_full_metrics_suppl} report the results under the in-dataset and the cross-dataset settings, respectively. Overall, it can be seen that LAA-Net achieves better performances than other state-of-the-art methods.

\subsection{Qualitative Results: E-FPN versus FPN}

A qualitative comparison between the proposed  E-FPN  and the traditional FPN with different fusion settings is reported in Figure~\ref{fig:efpn_fpn_abl}. Using EfficientNet-B4~\cite{efn_net} (EFNB4) as our backbone, the $\mathbf F^{(6)}$ refers to the features extracted from the last convolution block in the backbone. In other words, this means that no FPN design is integrated. By gradually aggregating features from lower to higher resolution layers, we can observe the improvement of the forgery localization ability for both E-FPN and FPN. More notably, E-FPN produces more precise activations on the blending boundaries as compared to FPN. This can be explained by the fact that the E-FPN integrates a filtering mechanism for learning less noise. In contrast, FPN seems to consider regions outside the blending boundary, which results in lower performance as previously shown in Table \textcolor{red}{4 - \textcolor{black}{Section} 4.4} of the main manuscript.

\subsection{Qualitative Results: Gaussian Noise}

\begin{figure}
    \centering
    \includegraphics[width=\linewidth]{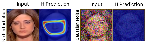}
    \caption{Detection of vulnerable points w/o and w/ Gaussian noise.}
    \vspace{-3mm}
    \label{fig:noise-refer}
\end{figure}

In Table 2 of the main manuscript, the performance of LAA-Net declined significantly when encountering Gaussian Noise perturbations. One possible reason is that the introduction of noise elevates the difficulty of detecting the vulnerable points. To confirm that, we report the inference of the heatmap before and after applying a Gaussian Noise on a facial image in Figure~\ref{fig:noise-refer}. As it can be observed, the detection of vulnerable points is highly impacted with the introduction of a Gaussian noise.

\subsection{Robustness to Compression}

To assess the robustness of LAA-Net to compression, we test LAA-Net on the c23 version of FF++, and the overall AUC is equal to $89.30$\%.